\documentclass[fleqn,utf8]{article}

\usepackage{PRIMEarxiv}

\usepackage[english]{babel}
\usepackage[utf8]{inputenc}
\usepackage[T1]{fontenc}
\usepackage{tabularx}
\usepackage{acronym}
\usepackage{bigstrut}
\usepackage{amsfonts}
\usepackage{amsmath}
\usepackage{algorithmic}
\usepackage{array}
\usepackage{color}
\usepackage{svg}
\usepackage{breakcites}
\usepackage{graphicx}
\usepackage{multirow}
\usepackage{bm}
\usepackage{url}
\usepackage{url,hyperref,lineno,microtype,subcaption}
\usepackage[onehalfspacing]{setspace}
\usepackage[square,numbers,sort&compress]{natbib}

\newcolumntype{?}{!{\vrule width 2.25pt}}
\newcolumntype{+}{!{\vrule width 1.125pt}}

\graphicspath{{Figures/}}

\title{Generalizing electrocardiogram delineation: training convolutional neural networks with synthetic data augmentation}

\author{
  Guillermo Jimenez-Perez, Juan Acosta \\
  Arrhythmia Unit, Department of Cardiology \\ Virgen Del Rocío University Hospital \\ \texttt{guillermo@jimenezperez.com}
  \And
  Alejandro Alcaine \\
  Facultad de Ciencias de la Salud \\ Universidad San Jorge
  \And
  Oscar Camara \\
  PhySense research group \\ Department of Information and \\ Communication Technologies \\ Pompeu Fabra University
}

\begin{document}
\maketitle
\vspace{-2em}

\begin{abstract}

Obtaining per-beat information is a key task in the analysis of cardiac electrocardiograms (ECG), as many downstream diagnosis tasks are dependent on ECG-based measurements. Those measurements, however, are costly to produce and time-consuming to process in bulk, especially in recordings that change throughout long periods of time. Currently, ECG delineation is performed either using digital signal processing (DSP), which are able to produce high-quality delineations but are difficult to generalize, and machine learning (ML), which commonly produces increased performance at the cost of needing large databases of annotated data. However, the existing annotated databases for ECG delineation are small, being insufficient in size and in the array of pathological conditions they represent. This article delves into the latter with two main contributions. First, a pseudo-synthetic data generation scheme was developed, based in probabilistically composing unseen ECG traces given ``pools'' of fundamental segments cropped from the original databases and a set of rules for their arrangement into coherent synthetic traces. The generation of conditions is controlled by imposing expert knowledge on the generated trace, which increases the input variability for training the model. Second, two novel segmentation-based loss functions have been developed, which attempt at enforcing the prediction of an exact number of independent structures and at producing closer segmentation boundaries by focusing on a reduced number of samples. The best performing model obtained an $F_1$-score of 99.38\% and a delineation error of $2.19 \pm 17.73$ ms and $4.45 \pm 18.32$ ms for all wave's fiducials (onsets and offsets, respectively), as averaged across the P, QRS and T waves for three distinct freely available databases. The excellent results were obtained despite the heterogeneous characteristics of the tested databases, in terms of lead configurations (Holter, 12-lead), sampling frequencies ($250$, $500$ and $2,000$ Hz) and represented pathophysiologies (e.g., different types of arrhythmias, sinus rhythm with structural heart disease), hinting at its generalization capabilities, while outperforming current state-of-the-art delineation approaches.

\end{abstract}

\clearpage

\section{Introduction}\label{sec:introduction}

The electrocardiogram (ECG) is one of the main measurement tools in clinical practice given its rich insight into the cardiac electrophysiology, its ease of use and its relative inexpensiveness compared to other diagnostic methods. The ECG reflects the electrical activity of the heart, which can be logically grouped as a set of waves corresponding to different phases in the cardiac cycle. Thus, the P wave corresponds to atrial depolarization, the QRS corresponds to ventricular depolarization and the T wave corresponds to ventricular repolarization \cite{macleod2013essential}. Extracting these waves (and their corresponding ST, PQ and TP segment pauses) allows the quantification of objective measurements of the heart's electrophysiological function \cite{macleod2013essential}, which can be used to characterize many pathological deviations from normal sinus rhythm (i.e., absence of P wave in ventricular rhythms or ST segment elevation/depression in myocardial infarction) \cite{macleod2013essential}. Moreover, these measurements are, in turn, employed in algorithms for diagnosis \cite{andreu2018qrs}, either as clinical thresholds that indicate deviations from normality or in machine learning algorithms as extracted features for training and testing models \cite{Minchole2019}, among others. In the case of the ECG, accurately and automatically measuring the different waves could aid in the development of more precise decision support systems or monitorization tools by aggregating information in multiple-lead registries for several heart cycles, which is a highly time-consuming task that hampers the cardiologists' workflow \cite{Minchole2019}.

Many computational approaches exist for the automatic quantification of the ECG. Most of these produce delineation of the electrocardiogram \cite{Martinez2004, Banerjee2012, Dubois2007, Graja2005, Camps2019, Jimenez-Perez2019, Jimenez-Perez2021, Hedayat2018, Tison2019, Sodmann2018}. Delineation methods can be divided in two main groups: digital signal processing (DSP) and machine learning (ML) based methods. The latter can be further subdivided into deep learning (DL) and non-DL (hereinafter ``hand-crafted'') methods.

Digital signal processing methods \cite{laguna1994automatic, Martinez2004, Banerjee2012, chen2020crucial} have the advantage of explicitly imposing priors on the biomarker extraction process, as they usually consist in a series of data-transformation steps (i.e., application of the wavelet transform) that reveal the cutoff points more clearly and a posterior rule-based algorithm for aggregating this partial information. These methods, however, often generalize poorly to unseen morphologies given their dependence on the production of robustly engineered transformation and rule-based aggregation steps \cite{Minchole2019}, thus becoming more difficult to maintain.

Machine learning methods, on the other hand, have different associated problems that hinder their widespread adoption. Hand-crafted ML algorithms \cite{Dubois2007, Graja2005} are difficult to train when using large amounts of annotated samples, which are becoming commonplace in current state-of-the-art, and usually provide reduced performance as compared to well-tuned DSP-based or DL-based solutions. The reason for this is that feature engineering, a key step in hand-crafted ML-based solutions, is costly and difficult to produce in a robust, fast and comprehensive manner \cite{Lyon2018}. DL-based methods \cite{Jimenez-Perez2019, Jimenez-Perez2021, Sodmann2018, Hedayat2018, Tison2019, moskalenko2019deep, sereda2019ecg}, on their behalf, provide black-box solutions that are difficult to verify, require large amounts of annotated data, have difficulties leveraging \textit{a priori} information and need quality loss functions for obtaining sensible data representations \cite{Minchole2019a,Minchole2019,kim2019learning,lecun2015deep}. Moreover, both hand-crafted- and DL-based algorithms face difficulties when learning ECG data, given its high beat-to-beat morphological similarity and the small size of current ECG databases for their usage in data-driven approaches.

For addressing the aforementioned issues with DL-based data analysis, a fully convolutional model was trained on the QT \cite{Laguna1997} and Lobachevsky University (LU) \cite{kalyakulina2020ludb} databases with a focus on direct data quantification, i.e., for producing ECG delineations. Using the U-Net \cite{Ronneberger2015} as the base architecture, some of the previously mentioned DL-based hindrances are addressed in three ways. Firstly, we developed a novel pseudo-synthetic data generation method for augmenting the database size with \textit{a priori} information on normal and pathological ECG behaviour. Secondly, two loss functions were developed: the BoundaryLoss, which provide enhanced pixel accuracy close to the segmentation borders and is similar to other approaches in the literature \cite{juhl2019guiding, cheng2021boundary}, and the F1InstanceLoss, which promotes cohesiveness in the predicted pixels regions. Lastly, we explored different modifications on the base architecture, namely different connectivity patterns such as the W-Net \cite{xia2017w,xu2020dw}, attention-based mechanisms \cite{wang2020eca} and different number of pooling operations. To the best of our knowledge, these improvements have not been explored in the literature for ECG analysis. A more rudimentary version of this work exists in the literature \cite{Jimenez-Perez2021}; however, the current approach displays key components that allow the algorithm to generalize better against a wider array of morphologies, such as the application of the synthetic data augmentation, the novel loss functions and a much wider array of architectural variation exploration.

The rest of the paper is organized as follows. Section \ref{sec:materials} describes the databases and methodology employed. Section \ref{sec:results} summarizes the main results. Finally, Section \ref{sec:discussion} discusses the obtained results in their context.

\section{Materials and Methods}\label{sec:materials}

This section firstly describes the used databases in Section \ref{sec:databases} for then defining the methodology employed for their analysis. The proposed methodology, on its behalf, can be divided into several steps. The first step consists in the pseudo-synthetic ECG generation from fundamental segments from a probabilistic rule-based algorithm (Section \ref{sec:synthetic}). The second step involve the definition and training of a deep learning architecture, which is subdivided into the description of the architecture itself (Section \ref{sec:architecture}) and the employed loss functions (Section \ref{sec:losses}). Finally, the evaluation metrics are described in Section \ref{sec:evaluation}. A final section was added for detailing the specific experiments performed (Section \ref{sec:experiments}). Our code will be made publicly available in \url{https://github.com/guillermo-jimenez/DelineatorSwitchAndCompose}.

\subsection{Databases}\label{sec:databases}

\begin{figure*}[!t]
    \centering
    \includegraphics[width=0.75\linewidth]{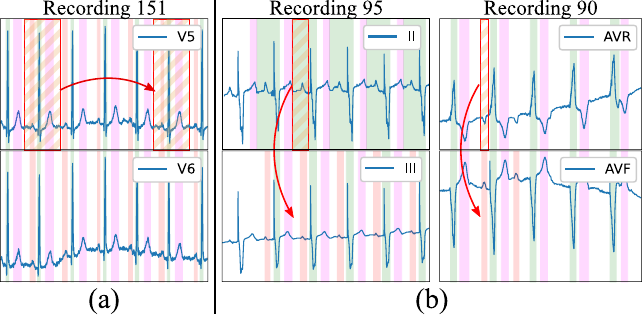}
    \caption{\label{fig:corrections} Limitations of existing delineation databases for training deep learning models. Examples in the LU database: a) High beat-to-beat redundancy within recordings; b) Incorrectly-annotated ground-truth (top lead) and correction (bottom lead). Color code: P wave (red), QRS wave (green) and T wave (magenta). Stripped segments highlight the errors. 
}
\end{figure*}

The QT \cite{Laguna1997}, the LU \cite{kalyakulina2020ludb} and the database from the Ningbo First Hospital of Zhejiang University \cite{zheng202012} (hereinafter the ``Zhejiang'' database) were employed for model training and evaluation. Specifically: the QT database was used for synthetic data generation and model training; the LU database, for synthetic data generation and evaluation; and the Zhejiang database, for model testing. The QT database contains $105$ two-lead ambulatory recordings of 15 minutes at $250$ Hz, representing different pathologies (arrhythmia, ischemic/non-ischemic ST episodes, slow ST level drift, transient ST depression and sudden cardiac death) as well as normal sinus rhythm. The LU database is composed of $200$ 12-lead recordings of 10 seconds of length, sampled at $500$ Hz, comprising sinus and abnormal rhythms as well as a variety of pathologies. The Zhejiang database, on its behalf, includes $334$ 12-lead OTVA recordings of variable size ($2.8 - 22.6$ seconds), sampled at $2,000$ Hz, and was originally devised for identifying the site of origin outflow tract ventricular arrhythmias (OTVAs), containing no delineation annotations. These databases are an appropriate sample for testing generalizability, since they present heterogeneity in their represented pathologies, sampling rates, lead configurations (Holter and standard 12-leads) and centers of acquisition.

The existing delineation databases have certain characteristics that hinder the development of reliable delineation algorithms. On the first hand, although they contain a relatively large amount of delineated cardiac cycles ($3,528$ and $1,830$ annotated beats in the QT and LU databases, respectively), these present a high intra- and inter-patient redundancy (i.e., very similar morphologies in different patients for certain pathologies or during sinus rhythm and very stable ECG beat-to-beat morphology in the same trace), which complicates model training due to reduced population variability (Figure \ref{fig:corrections}a). Moreover, given the difficulty and time-consuming process of delineating an ECG, some registries present delineation errors such as skipped beats or inconsistent onset/offset predictions for similar morphologies, among others. Those problems were addressed in two ways. Firstly, those outlier beats were reannotated when necessary with the help of an expert cardiologist. Secondly, new ground truth was generated for the Zhejiang database, which was not annotated for delineation purposes, and reserved for algorithm testing as an independent set. All these new annotations have been added as supplementary materials in the digital version. Some examples of annotation corrections can be seen in Figure \ref{fig:corrections}b.

\begin{figure*}[!p]
    \centering
    \includegraphics[width=1\linewidth]{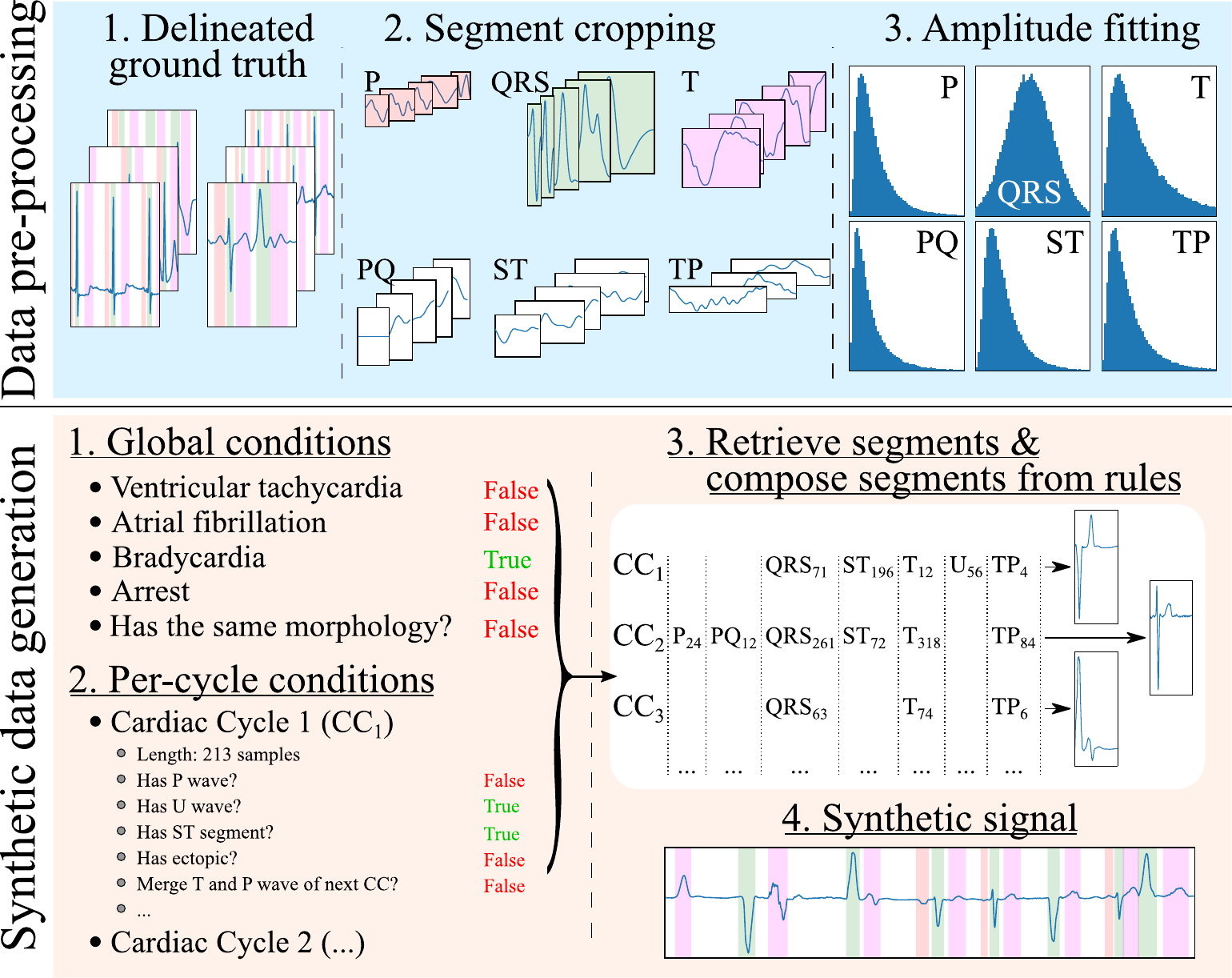}
    \caption{\label{fig:rulebased} Synthetic electrocardiogam generation pipeline. The data pre-processing step consists in: (1) delineating the ground truth; (2) cropping the different beats contained in the ground truth into their constituent segments (P, PQ, QRS, ST, T and TP), normalizing the QRS segment to have an absolute amplitude of 1, and normalizing the rest of the segments' as the amplitude fraction with respect to their (normalized) relative QRS; and (3) fitting the amplitudes to a normal distribution for the QRS wave (fraction with respect to its original amplitude) and log-normal distributions for the rest of the segments (fraction with respect to the QRS' amplitude). The synthetic data generation step, on its behalf, involves (1) producing a set of global rules that will be common for all synthesized cardiac cycles (in the example, the registry has bradycardia), (2) producing a set of rules that will affect each cardiac cycle individually (in the example, the first cardiac cycle, $\textrm{CC}_1$, skips its P wave to simulate a ventricularly-mediated beat or a very low amplitude P wave), (3) retrieving the specific segments and their amplitude relationships from the ``bags'' of cropped segments for their composition into independent cardiac cycles and (4) concatenating the segments into the final synthetic trace.
}
\end{figure*}

The data and ground truth, either real or synthesized, were then represented as binary masks for their usage in DL-based segmentation architectures, where a mask of shape $\{0,1\}^{3 \times N}$ was $True$-valued whenever a specific sample $n \in N$ was contained within a P, QRS or T wave (indices 0, 1 and 2, respectively) and $False$-valued otherwise \cite{Jimenez-Perez2021}, bridging the gap with the imaging literature. Finally, the joint training database was split into 5-fold cross-validation with strict subject-wise splitting, not sharing beats or leads of the same patient in the training and validation sets \cite{Faust2018, Jimenez-Perez2021}. Given that the proposed method employs pseudo-synthetic data generation, the pseudo-ECGs were also generated using data uniquely from the training set for each fold, ensuring no cross-fold contamination.

\subsection{Pseudo-Synthetic Data Generation}\label{sec:synthetic}

The structure of an ECG can be regarded as a combination of the P, QRS and T segments, alongside the PQ, ST and TP pause segments, which represent different phases of the electrical activation of the heart. The ECG is able to represent in its trace many pathological and non-pathological changes, reflecting slight deviations in its different constituting segments. The resulting ``modular'' structure can be leveraged in data-driven approaches for generating pseudo-synthetic data. 

The developed generation pipeline, depicted in Figure \ref{fig:rulebased}, consisted in two main stages: a pre-processing step that prepared the data for its posterior usage and a data generation step that created synthetic ECG traces through composing independently generated cardiac cycles. The data pre-processing step, on its behalf, involved cropping the delineated ground truth (in this case, the QT and LU databases) in its constituent segments and into separate ``pools'' of segments from which to draw in subsequent stages. Additionally, the segment's amplitude (relative to their associated QRS) was fitted into independent log-normal distributions, which would be sampled from in the generation step to relate the amplitude of each segment to the amplitude of the QRS in each cardiac cycle. The QRS segment amplitude was normalized with respect to the maximum QRS amplitude in the whole registry (comprising all leads).

The synthetic data generation step has several sub-steps. First, a set of global generation rules that affect all generated cardiac cycles were probabilistically generated for each sample. These have been limited to ventricular tachycardia (VT), atrial fibrillation (AF), atrioventricular (AV) blocks, sinus arrest (and its duration) and ST elevation/depression as a proof of concept. Second, a set of per-cardiac-cycle rules were generated, such as the presence or absence of each specific segment (P, QRS+T, PQ, ST, TP and U), whether the cycle corresponded to a ventricular ectopic (larger QRS amplitude and duration, absence of P wave) or whether there was wave merging (P with QRS, QRS with T, T with the next cycle's P). In the first and second steps, the rules were defined by drawing samples from a uniform distribution and applying the associated operation (global in the first case and per-cycle in the latter) in case they surpassed a pre-defined threshold.

Third, a set of segments were randomly selected from the segment ``pools''. A set of operations were then applied when extracting the segments from the pools as well as on the resulting cardiac cycles to comply with the global and per-cycle conditions. In particular, these operations comprised setting the segment's amplitude, interpolating the segment to a randomized number of samples to enforce as much variability as possible, cropping the segment, merging of the segment with the next (e.g., merging the T and the P waves, thus enforcing TP segment suppression), sign-correcting the segment to match other cardiac cycles or applying per-segment elevation/depression. 

Finally, the final synthetic signal and the ground truth were composed from the individual cardiac cycles. A set of post-composition operations were added to further increase the generated signal's variability, consisting in adding baseline wander noise, interpolating to slower or faster rhythms, adding flat line noise at the signal's edge, setting the global amplitude (multiplying the amplitude by a factor) and defining the trace's starting segment.

\begin{figure*}[t!]
    \centering
    \includegraphics[width=1\linewidth]{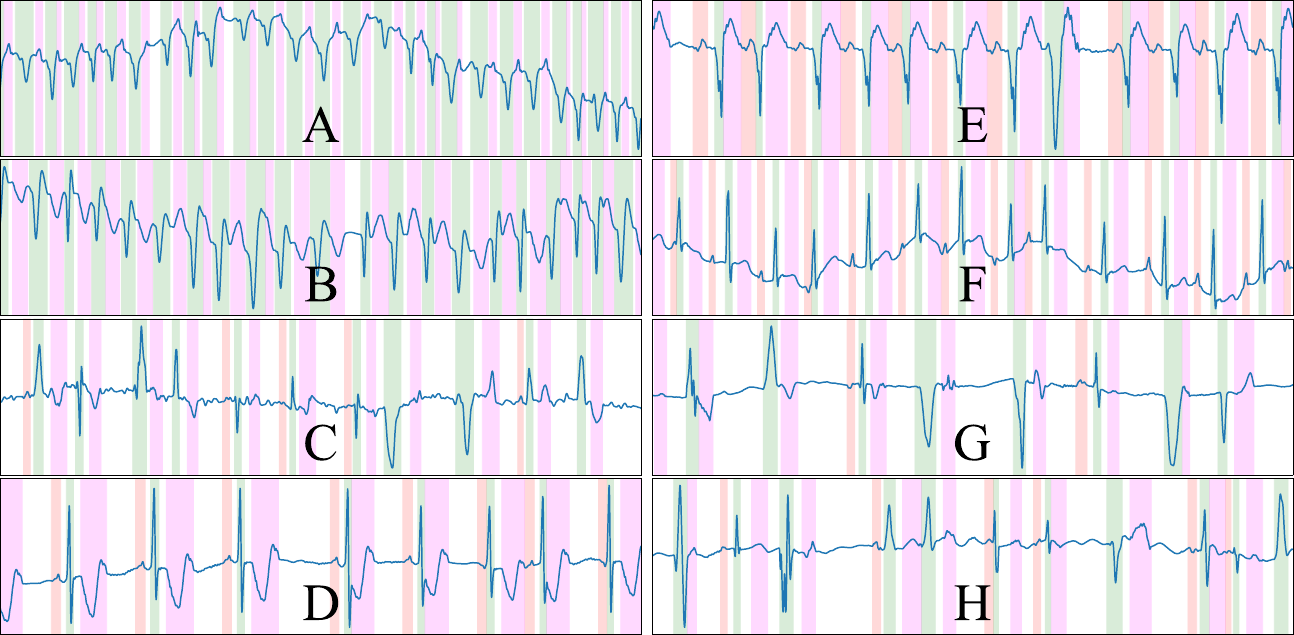}
    \caption{\label{fig:synthetic} Randomly drawn samples from the developed synthetic data generator. The generator is able to produce samples of a variety of conditions such as ventricular tachycardia or atrial fibrillation, among others. The samples presented display ventricular ectopics (C, G, H), sinus rhythm (D, E, F), atrial fibrillation (C) and ventricular tachycardia (A, B), and are generated alongside their ground truth. Color code: P wave (red), QRS wave (green) and T wave (magenta).
}
\end{figure*}

An important aspect to pseudo-synthetic ECG generation is efficiency, as the samples were generated online rather than offline to avoid restricting the approach to a fixed set of previously drawn samples. This is, however, only relevant during the training phases of the model, but can limit the options of operations that can be performed on the algorithm; in fact, many of the chosen additions were limited in their scope by this constraint, being restricted sometimes to oversimplified operations that offer close-enough approximations of the underlying represented cardiac conditions. Some randomly drawn samples from the synthetic data generator are shown in Figure \ref{fig:synthetic}.

\subsection{Architecture}\label{sec:architecture}

\begin{figure*}[t!]
    \centering
    \includegraphics[width=1\linewidth]{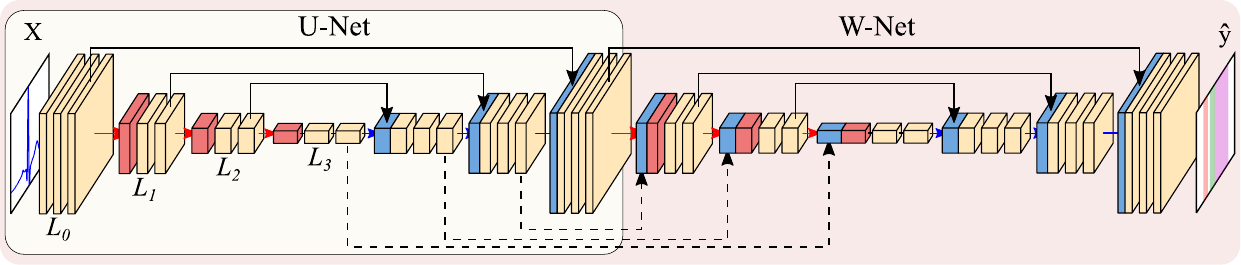}
    \caption{\label{fig:architecture} Representation of the U-Net (encircled in yellow) and W-Net architectures (encircled in red, containing the U-Net). Both networks are instantiated with 3 levels and 2 convolutional blocks per level. Arrows represent operations, while blocks are indicative of output tensors. Convolutional filters are doubled at each level, so that level $L_i$ has $2^\textit{i} N$ channels per level (with N being the starting number of channels), whereas pooling and upsampling have a kernel size of $2$. Color code: convolutions (yellow), pooling operations (red), upsampling operations (blue), concatenation operations (black).
}
\end{figure*}

The U-Net \cite{Ronneberger2015} is a convolutional neural network is an encoder-decoder structure, as depicted in Figure \ref{fig:architecture}. The encoder extracts high-level representations of the input data by means of convolutional operations, which transform an input tensor by convolving it with a trainable kernel, and pooling operations, which allow for reducing computational complexity. The decoder, on the other hand, upsamples the high-level encoder tensor to recover the original input's resolution while aggregating partial results obtained in different levels of the encoder. This direct feature aggregation between the encoder and the decoder, in the shape of tensor concatenation, allows for finer border definitions while avoiding gradient vanishing problems \cite{Ronneberger2015}. As in the original article, the number of trainable convolutional filters is doubled after every pooling operation and halved after every upsampling operation.

Many authors have experimented with the hyperparameters governing the U-Net, in the shape of number of convolutional operations (width), the number of upsampling-downsampling pairs (depth), starting number of convolutional filters, type of convolutional operation, type of non-linearity and presence/absence of other post-convolutional operations (batch normalization \cite{ioffe2015batch}, spatial dropout \cite{Tompson2015}), among others, which was partially covered in \cite{Jimenez-Perez2021} for the QT database.

Other authors have explored refining further architectural changes. Given the myriad of options, we restricted the exploration to the application of the W-Net architecture due to its good performance in other segmentation domains \cite{xu2020convolutional} as well as the usage of self-attention mechanisms in the shape of efficient channel attention (ECA). The W-Net \cite{xia2017w,xu2020dw} involves the application of a second U-Net whose input is the output of the first U-Net, thus approximately doubling the amount of parameters for the same number of initial channels. The W-Net also concatenates the tensors at the decoder of the first U-Net with the encoder of the second, similarly to the connections established between the encoder and the decoder of a ``vanilla'' U-Net. This secondary structure makes the network deeper, which usually presents increased performance \cite{Szegedy2014}. Self-attention applies the attention mechanism to a tensor, thus allowing different elements of the tensor to evaluate their relative importance for obtaining a certain result. This usually improves overall model performance and explainability \cite{prabhakararao2020myocardial}. ECA, specifically, is an approach to apply this mechanism to CNNs in an efficient manner \cite{wang2020eca}.

\subsection{Loss functions}\label{sec:losses}

Two novel loss functions, the BoundaryLoss and the F1InstanceLoss, were developed with the objective of enhancing the resulting prediction accuracy in two ways: the F1InstanceLoss enforces the retrieval of connected structures so that a penalty term is induced if the number of predicted and present structures differ; the BoundaryLoss attempts at adapting more tightly to the target boundary by means of computing the intersection-over-union of a subset of the original samples present in a mask, as opposed to the usual Dice score computation. These losses were based on the application of edge detectors, embedded into convolutional operations (flagged as non-trainable and with fixed weights), allowing automatic differentiation for posterior gradient propagation.

The first step consisted in applying the edge detector along all non-batch and non-channel axes of the input tensors, isolating the segmentation boundary. In the case of the BoundaryLoss, a large kernel size is employed ($K \in \mathbb{R}^{n}$, $n$ being an hyperparameter), whereas in the F1InstanceLoss the kernel size remains small ($K \in \mathbb{R}^{3}$). In this case the Prewitt operator was employed as the edge detector, which is defined as:

\begin{equation}
    \hspace{3em}\Vec{K}_{F_1} = \left(\begin{matrix} -1 & 0 & +1 \end{matrix}\right)^T,\hspace{10em}\Vec{K}_{Bound} = \left(\begin{matrix} -1 & 0 & ... & 0 & +1 \end{matrix}\right)^T
\end{equation}

The second step took the absolute of the edge-detected tensors for both the predicted and the ground truth masks. In the case of the BoundaryLoss, the third step involved the calculation of the Dice coefficient between the resulting tensors. This has the advantage of comparing the mask overlap on a reduced pool of pixels, increasing the precision at the segmentation boundary, as is the case in usual image processing pipelines. In the case of the F1InstanceLoss, the third step was based on summing the border activations along each non-batch and non-channel axis separately for both the predicted and ground truth tensors, obtaining the number of discontinuities present in the binary mask. These discontinuities act as surrogates of the onset/offset pairs of the binary masks, thus allowing the computation of  number of predicted and ground truth elements ($P_{elem}$ and $GT_{elem}$, respectively) for their usage in usual precision and recall metrics in a fully differentiable manner. The true positive (TP), false positive (FP) and false negative (FN) metrics are then computed by clamping these values, so that:

\begin{equation}
    \begin{array}{lll}
        TP & = & \left|GT_{elem}-max(GT_{elem}-P_{elem},\,0)\right| \\  
        FP & = & max(P_{elem}-GT_{elem},\,0) \\  
        FN & = & max(GT_{elem}-P_{elem},\,0) \\  
    \end{array}
\end{equation}

\begin{figure*}[!t]
    \centering
    \includegraphics[width=1\linewidth]{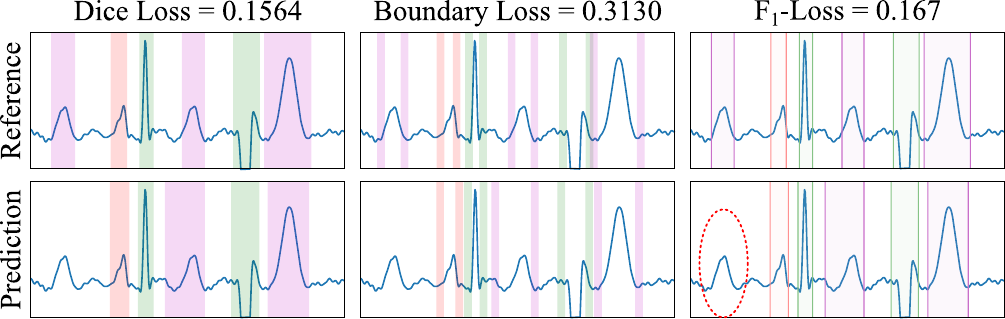}
    \caption{\label{fig:lossfunctions} Example of loss functions applied to a sample from the LU database. The Dice loss measures the overlap between the ground truth mask (GT, top) and the predicted mask (bottom). The BoundaryLoss computes a secondary mask for isolating samples surrounding the boundaries of the GT and predictions, thus more specifically penalizing onsets/offset mistakes. The $F_1$-InstanceLoss locates the onsets/offsets pairs of the masks for using these as surrogates of precision and recall metrics. In the example, the ground truth contains three T waves, whereas only four T waves have been predicted; the $F_1$-score loss for each individual wave is, thus, $0$, $0$ and $0.167$, respectively, given the false negative in the T wave. Color code: P wave (red), QRS wave (green) and T wave (magenta).
}
\end{figure*}

Finally, the TP, FP and FN values are then used to compute the smoothed $F_1$-score between the input and target masks. The computation process of these loss functions is depicted in Figure \ref{fig:lossfunctions}.

\subsection{Evaluation}\label{sec:evaluation}

The model evaluation is based on the computation of detection metrics, i.e., the model's precision, recall and $F_1$-Score, and delineation metrics, i.e., onset and offset errors on the true positives (mean, M $\pm$ standard deviation, SD). The computation of the metrics consisted in three steps. Firstly, the onset and offset fiducials were retrieved from the predicted binary mask (described in Section \ref{sec:databases}) to express the sample of occurrence by retrieving the locations of value change ($False$ to $True$ or vice-versa). Secondly, the ground truth and predicted fiducials were matched through a correspondence matrix. Thirdly, the correspondence matrix was used to compute the detection and delineation metrics.

The correspondence matrices between the true ($P$, $QRS$ and $T$) and predicted ($\hat{P}$, $\hat{QRS}$ and $\hat{T}$) fiducials were computed as:

\begin{equation}\label{eqn:corrmatrix}\arraycolsep=1.4pt
    \begin{array}{ll}
        P_{ij} = 
        \left\lbrace
        \begin{array}{lll} 
            1 & \textrm{if} & \hat{P}_{\textrm{fid}}[j] \in [P_{\textrm{on}}[i], P_{\textrm{off}}[i]] \\
            & \textrm{or} & P_{\textrm{fid}}[j] \in [\hat{P}_{\textrm{on}}[i], \hat{P}_{\textrm{off}}[i]] \\
            0 & \multicolumn{2}{l}{\textrm{otherwise}}
        \end{array}
        \right. & \hspace{2em} QRS_{ij} = 
        \left\lbrace
        \begin{array}{lll} 
            1 & \textrm{if} & \hat{QRS}_{\textrm{fid}}[j] \in [QRS_{\textrm{on}}[i], QRS_{\textrm{off}}[i]] \\
            & \textrm{or} & QRS_{\textrm{fid}}[j] \in [\hat{QRS}_{\textrm{on}}[i], \hat{QRS}_{\textrm{off}}[i]] \\
            0 & \multicolumn{2}{l}{\textrm{otherwise}}
        \end{array}
        \right. \\ \\
        T_{ij} = 
        \left\lbrace
        \begin{array}{lll} 
            1 & \textrm{if} & \hat{T}_{\textrm{fid}}[j] \in [T_{\textrm{on}}[i], T_{\textrm{off}}[i]] \\
            & \textrm{or} & T_{\textrm{fid}}[j] \in [\hat{T}_{\textrm{on}}[i], \hat{T}_{\textrm{off}}[i]] \\
            0 & \multicolumn{2}{l}{\textrm{otherwise}}
        \end{array}
        \right. &
    \end{array}
\end{equation}

where $\textrm{fid} \in \{\textrm{on, peak, off}\}$ is the specific fiducial to be explored, and $i \in [0,M]$, $j \in [0,N]$ are the total true and predicted fiducials for each of the waves, respectively.

These correspondence matrices were used to obtain the detection and delineation metrics. The detection metrics (true positives, TP, false positives, FP, and false negatives, FN) were computed as follows: given a correspondence matrix $H$, true positives were computed as elements that have been matched ($\textrm{TP} = \sum H_{ij}$); false positives were elements of a predicted fiducial that did not match any element in the ground truth, corresponding to the difference between the number of predicted fiducials and the cardinality of the matches ($\textrm{FP} = N - card(\{(i,j) \mid H_{ij} = 1 \})$); and false negatives were computed as elements of the ground truth that did not match any true fiducial, corresponding to the difference between the number of true fiducials and the cardinality of the matches ($\textrm{FP} = M - card(\{(i,j) \mid H_{ij} = 1 \})$). The TP, FP and FN were in turn used to compute the model's precision ($Pr$), recall ($Re$) and $F_1$-score. The delineation error, on its behalf, was computed through the mean and standard deviation (SD) of the difference of the actual and predicted onsets and offsets of the TP in the correspondence matrix:

\begin{equation}
    \min_{i, j} \;\; w_{\textrm{fid}}[i] - \tilde{w}_{\textrm{fid}}[j] \;\;\;\;\;\;\; \textrm{s.t.} \; H_{ij} = 1.
\end{equation}

These metrics were employed in turn for assessing the performance on the QT, LU and Zhejiang databases. In the case of the QT database, to homogenize the evaluation criteria with the existing literature, the detection and delineation metrics were computed for single-lead and multi-lead approaches, where the single-lead is based on evaluating the performance of both leads in the Holter registry independently, and the multi-lead consists in taking, for each beat, the lead that produces the best adjustment. Contrarily, the LU and Zhejiang databases were evaluated by fusing the individual lead predictions to obtain a single output prediction for, posteriorly, comparing this delineation with the annotated ground truth. The final prediction was computed through combining the individual lead results using majority voting of the 12 leads and the different models resulting from training on separate folds of the QT database, forming an ensemble.

Finally, these metrics were also used to define the ``best'' performing model, which was selected as the one producing good detection performance while attaining the lowest possible delineation error for the QT (in the validation fold), LU and Zhejiang databases. This was addressed through the calculation of two figures of merit: the largest $F_1$-score as detection performance and the smallest SD of the error as delineation performance for all three databases across all waves, and reported in Section \ref{sec:bestarch}. Moreover, this model ranking was employed for producing ablations of the different modifications (Section \ref{sec:materials}) by isolating a single modified factor while leaving the rest of the hyperparameters unmodified. These have been reported in Section \ref{sec:bestadditions}.

\subsection{Experiments}\label{sec:experiments}

The model's performance was tested under an array of complementary tests to address the contributions of the different elements to the results. Firstly, the importance of the pseudo-synthetic data generation was addressed by training the same model architecture using augmented data (real and synthetic), synthetic-only data and real-only data. Identical computational budget was ensured by producing the same number of batches (with identical batch size) for the same number of epochs by oversampling the training database. Secondly, the importance of the BoundaryLoss and F1InstanceLoss was addressed also by doubling the number of executions, with and without the proposed losses. The Dice score always remained as a baseline for training in every configuration. Finally, the importance of the architectural modifications was addressed. Several architectures were tested: U-Net for depths 5, 6 and 7; W-Net for depths 5 and 6; and W-Net with ECA for depth 5. In all cases, the number of input channels was kept the same in the W-Net as in its U-Net counterpart, resulting in models with increased number of parameters (capacity). These were selected to have as many candidate architectures as possible but without compromising the computational budget of our equipment. In total, 66 different configurations were tested to address the model's performance.

Some design choices were kept constant to avoid unfeasibly large hyperparameter exploration. All model configurations used the same random seed (123456), leaky ReLU non-linearities, zero padding for preserving tensor shape, kernels of size 3, batch normalization, spatial dropout \cite{Tompson2015} ($p = 0.25$), Adam optimizer \cite{Kingma2015} ($lr = 0.001$) and the Dice loss alongside the developed losses. The BoundaryLoss employed a kernel size of 11 samples. The ordering of operations after the convolutional operations was defined to agree with the image segmentation literature (non-linearity $\rightarrow$ batch normalization $\rightarrow$ dropout) \cite{Chollet2017, He2016b}. All networks were trained using ECG-centered data augmentation, as described elsewhere \cite{Jimenez-Perez2021}, comprising additive white Gaussian noise, random periodic spikes, amplifier saturation, powerline noise, baseline wander and pacemaker spikes to enhance the model's generalizability. All executions were performed at the Universitat Pompeu Fabra's high performance computing environment, assigning the jobs to either an NVIDIA 1080Ti or NVIDIA Titan Xp GPU, and used the PyTorch library \cite{paszke2019pytorch}.

\section{Results}\label{sec:results}

\begin{table*}[t!]
\centering
{
\begin{tabular}{ll|c|c?c|c|c|c|}
\cline{3-8}
 &  & \begin{tabular}[c]{@{}c@{}}This work \\ (SL)\end{tabular} & \begin{tabular}[c]{@{}c@{}}This work \\ (ML)\end{tabular} & \cite{Jimenez-Perez2021}, SL & \cite{Jimenez-Perez2021}, ML & \cite{Martinez2004} & \cite{Camps2019} \\ \hline
\multicolumn{1}{|l|}{\multirow{4}{*}{\rotatebox[origin=c]{90}{P wave}}} & Pr & \textbf{99.27} & 98.90 & 90.12 & 94.17 & 91.03 & N/R \\ \cline{2-8} 
\multicolumn{1}{|l|}{} & Re & 98.38 & \textbf{99.72} & 98.73 & 94.70 & 98.87 & N/R \\ \cline{2-8} 
\multicolumn{1}{|l|}{} & OnE & -1.2 $\pm$ 17.9 & \textbf{-0.8 $\pm$ 13.5} & 1.5 $\pm$ 22.9 & -1.7 $\pm$ 17.8 & 2.0 $\pm$ 14.8 & N/R \\ \cline{2-8} 
\multicolumn{1}{|l|}{} & OffE &  1.1 $\pm$ 16.6 & \textbf{-0.6 $\pm$ 12.7} & 0.3 $\pm$ 16.0 & 4.0 $\pm$ 16.1 & 1.9 $\pm$ 12.8 & N/R \\ \hline \hline
\multicolumn{1}{|l|}{\multirow{4}{*}{\rotatebox[origin=c]{90}{QRS wave}}} & Pr & 99.31 & 99.24 & 99.14 & 99.40 & \textbf{99.86} & N/R \\ \cline{2-8} 
\multicolumn{1}{|l|}{} & Re & 99.94 & \textbf{99.97} & 99.94 & 99.28 & 99.80 & N/R \\ \cline{2-8} 
\multicolumn{1}{|l|}{} & OnE & -0.5 $\pm$ 11.2 &  \textbf{0.1 $\pm$  7.5} & -0.1 $\pm$ 8.4 & -3.8 $\pm$ 14.6 & 4.6 $\pm$ 7.7 & -2.6 $\pm$ 10.8 \\ \cline{2-8} 
\multicolumn{1}{|l|}{} & OffE &  3.7 $\pm$ 13.1 &  \textbf{1.7 $\pm$  7.8} & 3.6 $\pm$ 12.6 & 5.4 $\pm$ 16.8 & 0.8 $\pm$ 8.7 & 4.4 $\pm$ 15.2 \\ \hline \hline
\multicolumn{1}{|l|}{\multirow{4}{*}{\rotatebox[origin=c]{90}{T wave}}} & Pr & \textbf{98.73} & 98.24 & 98.25 & 96.36 & 97.79 & N/R \\ \cline{2-8} 
\multicolumn{1}{|l|}{} & Re & 99.78 & \textbf{99.97} & 99.88 & 99.09 & 99.77 & N/R \\ \cline{2-8} 
\multicolumn{1}{|l|}{} & OnE &  5.8 $\pm$ 39.6 &  \textbf{5.2 $\pm$ 31.1} & 21.6 $\pm$ 66.3 & 19.1 $\pm$ 66.5 & N/R & N/R \\ \cline{2-8} 
\multicolumn{1}{|l|}{} & OffE &  2.4 $\pm$ 51.3 &  3.8 $\pm$ 37.2 & 4.6 $\pm$ 31.1 & 9.9 $\pm$ 46.3 & \textbf{-1.6 $\pm$ 18.1} & N/R \\ \hline
\end{tabular}}
\caption{Precision (Pr, \%), recall (Re, \%), onset error (OnE, mean [M] $\pm$ standard deviation [SD], in miliseconds) and offset errors (OffE, M $\pm$ SD, in miliseconds) of our best performing single-lead (SL) and multi-lead (ML) models in the QT database. N/R stands for ``not reported''.}
\label{tab:resultsbest}
\end{table*}

\begin{table*}[t!]
\centering
{
\begin{tabular}{ll|c?c|c|c|c|c|}
\cline{3-8}
 &  & \begin{tabular}[c]{@{}c@{}}\textbf{Zhejiang} \\ (this work)\end{tabular} & \begin{tabular}[c]{@{}c@{}}\textbf{LU} \\ (this work)\end{tabular} & LU (\cite{chen2020crucial}) & \begin{tabular}[c]{@{}c@{}}LU (\cite{laguna1994automatic},\\ \textit{via} \cite{chen2020crucial})\end{tabular} & LU (\cite{moskalenko2019deep}) & LU (\cite{sereda2019ecg}) \\ \hline
\multicolumn{1}{|l|}{\multirow{4}{*}{\rotatebox[origin=c]{90}{P wave}}} & Pr & \textbf{97.57} & \textbf{99.62} & 98.43 & 98.43 & 97.69 & 90.48 \\ \cline{2-8} 
\multicolumn{1}{|l|}{} & Re & \textbf{98.65} & \textbf{99.81} & 96.44 & 96.44 & 98.01 & 97.36 \\ \cline{2-8} 
\multicolumn{1}{|l|}{} & OnE & \textbf{2.46 $\pm$ 12.58} & 8.23$\pm$ 9.01 & \textbf{2.2 $\pm$ 7.4} & 2.8 $\pm$ 7.5 & -0.6 $\pm$ 17.5 & 3.4 $\pm$ 18.4 \\ \cline{2-8} 
\multicolumn{1}{|l|}{} & OffE & \textbf{2.87 $\pm$ 12.43} & 3.01$\pm$10.40 & -6.5 $\pm$ 10.7 & \textbf{-7.3 $\pm$ 10.1} & -2.4 $\pm$ 18.4 & -4.1 $\pm$ 19.4 \\ \hline \hline
\multicolumn{1}{|l|}{\multirow{4}{*}{\rotatebox[origin=c]{90}{QRS wave}}} & Pr & \textbf{99.53} & \textbf{100.00} & 100.0 & 99.56 & 99.93 & 98.27 \\ \cline{2-8} 
\multicolumn{1}{|l|}{} & Re & \textbf{99.87} & \textbf{100.00} & 99.86 & 99.86 & 100.0 & 99.86 \\ \cline{2-8} 
\multicolumn{1}{|l|}{} & OnE & \textbf{4.72 $\pm$ 13.35} & \textbf{4.27$\pm$ 9.75} & 15.4 $\pm$ 14.6 & 18.4 $\pm$ 14.7 & 1.5 $\pm$ 11.1 & 1.7 $\pm$ 10.0 \\ \cline{2-8} 
\multicolumn{1}{|l|}{} & OffE & \textbf{3.26 $\pm$ 11.91} & \textbf{4.00$\pm$ 9.14} & -3.8 $\pm$ 13.6 & -5.4 $\pm$ 14.3 & 2.0 $\pm$ 10.6 & -3.4 $\pm$ 12.3 \\ \hline \hline
\multicolumn{1}{|l|}{\multirow{4}{*}{\rotatebox[origin=c]{90}{T wave}}} & Pr & \textbf{98.86} & \textbf{100.00} & 99.21 & 99.09 & 99.37 & 96.23 \\ \cline{2-8} 
\multicolumn{1}{|l|}{} & Re & \textbf{99.86} & 1\textbf{00.00} & 98.85 & 98.85 & 99.68 & 93.51 \\ \cline{2-8} 
\multicolumn{1}{|l|}{} & OnE & \textbf{8.73 $\pm$ 28.85} & 18.26$\pm$18.21 & \textbf{-1.3 $\pm$ 8.8} & -2.6 $\pm$ 11.4 & 2.9 $\pm$ 23.7 & 9.2 $\pm$ 28.2 \\ \cline{2-8} 
\multicolumn{1}{|l|}{} & OffE & \textbf{-3.77 $\pm$ 24.32} & -8.84$\pm$18.05 & \textbf{-1.2 $\pm$ 6.8} & -3.3 $\pm$ 7.3 & -2.4 $\pm$ 30.4 & -6.0 $\pm$ 25.0 \\ \hline
\end{tabular}
}
\caption{\label{tab:resultsluzhejiang}Precision (Pr, \%), recall (Re, \%), onset error (OnE, mean [M] $\pm$ standard deviation [SD], in miliseconds) and offset errors (OffE, M $\pm$ SD, in miliseconds) of our best performing model in the LU and Zhejiang databases, obtained through pixel-wise majority voting of the model developed for each fold trained on the QT database.}
\end{table*}

\subsection{Best performing model}\label{sec:bestarch}

The best performing model according to the criteria presented in Section \ref{sec:evaluation} was a self-attention W-Net model with 5 levels, trained with both real and synthetic data, while excluding the F1InstanceLoss and the BoundaryLoss. The model obtained an average $F_1$-score of 99.38\% and a average delineation error of $2.19 \pm 17.73$ ms and $4.45 \pm 18.32$ ms for the onsets and offsets, respectively, across all waves and databases. The per-database and per-wave metrics of the model (precision, recall, onset error and offset error) were reported in Tables \ref{tab:resultsbest} and \ref{tab:resultsluzhejiang} for completeness.

\subsection{Performance comparison of model additions}\label{sec:bestadditions}

\begin{figure*}[!p]
    \centering
    \caption*{\textbf{Training data source}}
    \vspace{-1em}
    \includegraphics[width=0.9\linewidth]{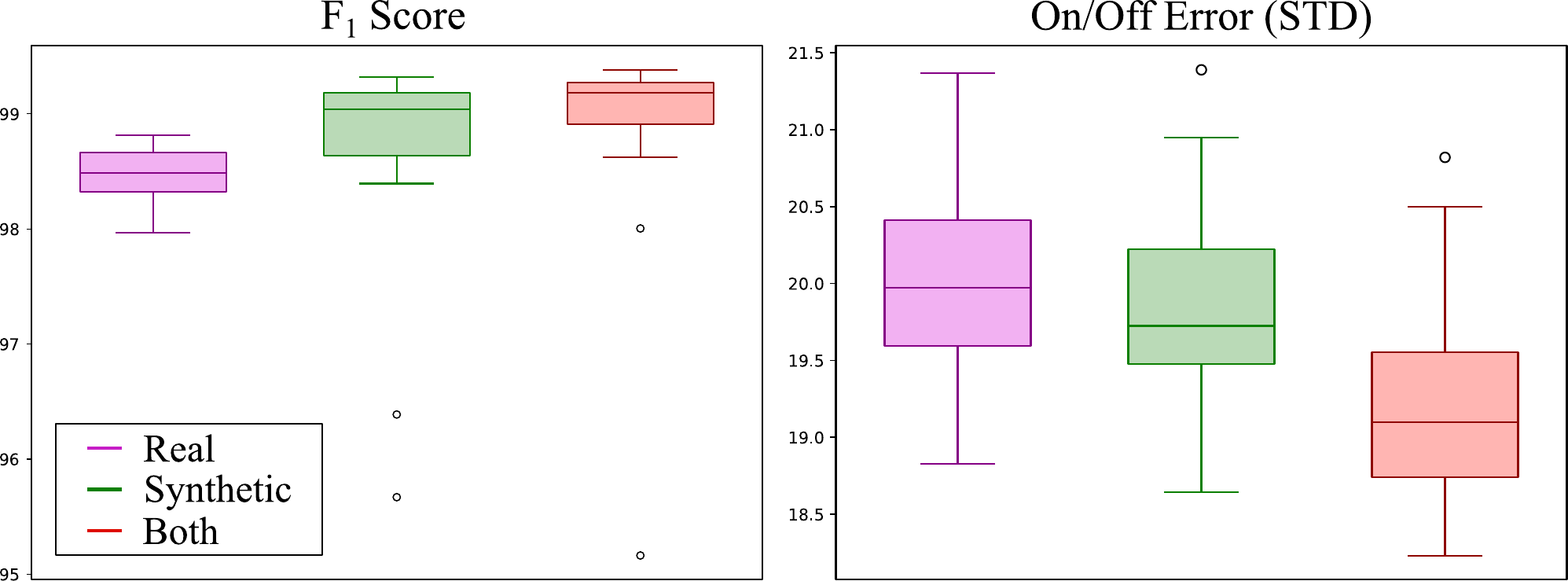}
    \vspace{1em}
    \caption*{\textbf{Model topology}}
    \vspace{-1em}
    \includegraphics[width=0.9\linewidth]{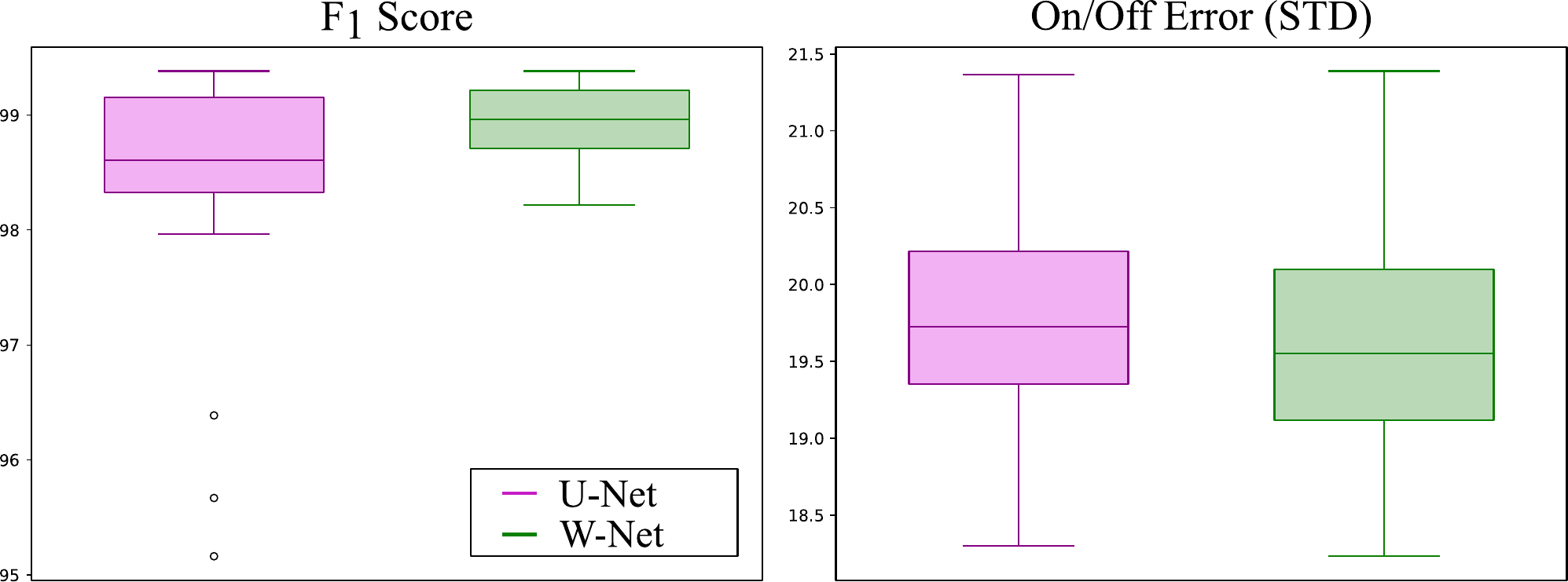}
    \vspace{1em}
    \caption*{\textbf{Loss function}}
    \vspace{-1em}
    \includegraphics[width=0.9\linewidth]{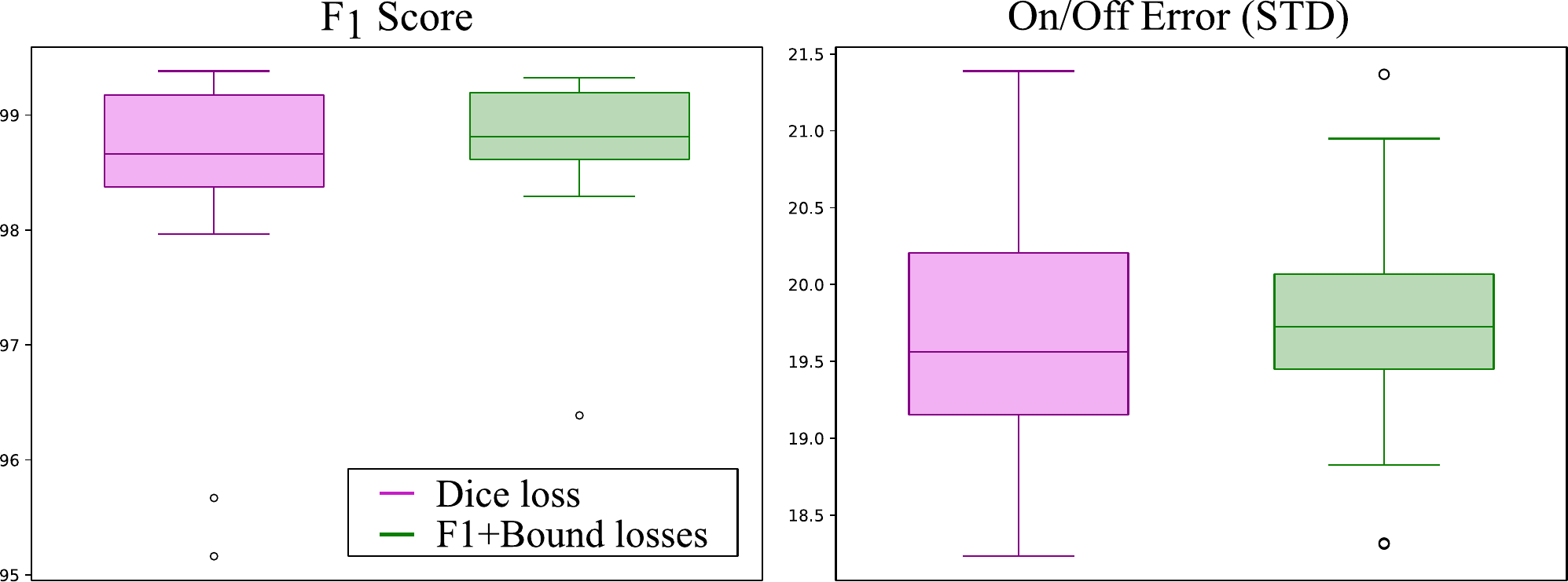}
    \vspace{1em}
    \caption{\label{fig:modeladditions} Detection (left; higher is better) and delineation (right; lower is better) performance of all models grouped by training data source, model topology and employed loss function. Synthetic-only data (green) showed higher detection and delineation performance than real-only data (magenta), whereas using both sources produced the best results for both detection and delineation performance. The W-Net (green) showed slightly higher detection and delineation performance than the U-Net (magenta). Finally, using the $F_1$-InstanceLoss and BoundaryLoss (green) resulted in models with higher detection performance but slightly lower delineation performance as compared to using Dice loss only (magenta).
}
\end{figure*}

The best performing additions were the usage of synthetic data generation, where the usage of both real and synthetic data reported an increased performance of 0.62\% on average with respect to the usage of real-only data. Interestingly, using synthetic data only for model training still produced increased performance over using real-only data, surpassing its performance by 0.35\% on average. Boxplots of the models grouped by data source can be visualized in Figure \ref{fig:modeladditions} (top). 

The second-to-best model performance addition was the usage of W-Net, which produced 0.53\% less delineation error and a reduction in its SD of $1.83$ ms and $2.20$ ms for the onset and offset metrics as compared to its U-Net counterpart (see Figure \ref{fig:modeladditions}, center). The third best model addition was the inclusion of the F1InstanceLoss and the BoundaryLoss functions, with added predictive performance of 0.26\% $F_1$-score as well as reducing the offset error in $0.07$ ms but increasing the onset error in $0.17$ ms (Figure \ref{fig:modeladditions}, bottom). The rest of the improvements (usage of self-attention, increase of model capacity) did not show a consistent effect on model performance.

\section{Discussion}\label{sec:discussion}

Deep learning commonly displays improved performance over many hand-crafted data analysis methodologies for a wide array of tasks \cite{tajbakhsh2020embracing}, due to their ability to leverage large pools of data, their adaptability to a wide range of tasks, the built-in feature engineering and the availability of of open code and large size datasets \cite{lecun2015deep}. However, DL algorithms have a series of drawbacks that hinder their implementation in data-sensitive contexts. Firstly, they have a large dependence on the size of the training data \cite{Minchole2019}, which might be difficult or expensive to acquire and annotate in many contexts such as in clinical environments. Secondly, DL models find difficulties when leveraging data priors, i.e., information that the system's designer knows must be implemented in the system. Some examples for the ECG would be the P wave (which might have imperceptible amplitude or might be masked within a QRS complex) or the fact that no T wave can exist without a QRS complex. Finally, the black-box nature of the models, where most works cannot guarantee that their predictions are not the result from spurious or acausal relations within the data or data leakage \cite{lazer2014parable}.

Some solutions exist for addressing these issues. Data scarcity has been addressed in the literature through pseudo labels \cite{pham2020meta} or through synthetic data generation, either using simulations \cite{doste2020silico} or generative adversarial networks (GANs), but these present efficiency issues (simulations) or face difficulties when extending beyond the training data manifold (data-driven approaches). Data priors, on their behalf, have been enforced either by producing representations that explicitly exclude previously known information via minimizing mutual information \cite{kim2019learning} or by providing the specific prior as input data (e.g., by including the label as an input to the model, such as in conditional GANs \cite{mirza2014conditional}), but the ability to explicitly control data-side priors is still limited. Finally, some authors have attempted to enhance model explainability through filter visualization methods \cite{selvaraju2017grad} or attention maps \cite{prabhakararao2020myocardial}, but these can only act as surrogates of decision rules and need interpreting on their own. A middle ground for DL on data-sensitive contexts is to employ DL for quantification tasks, such as segmentation. These intermediate networks do not provide an instantaneous prediction (e.g., diagnosis), but present the advantage of being immediately interpretable by an operator \cite{Minchole2019a}, avoiding placing confidence in faulty predictions while enabling many downstream tasks \cite{Lyon2018}. This, however, contrasts with the current scientific production; as an example, in the specific case of DL-based analysis on the ECG, the vast majority of published works focused on classification. Some authors have data scarcity, expensiveness of data annotation and lack of generalizability of existing solutions \cite{Minchole2019,Minchole2019a}.

This work addresses these issues by producing a network for ECG segmentation. Given the small size of these databases, the models were enriched with a novel pseudo-synthetic data generation strategy, which allowed for imposing expert knowledge through constraining the topology of the generated data. These priors were further enforced in the shape of two novel loss functions by minimizing the boundary error with respect to the reference (BoundaryLoss) and by maximizing precision and recall metrics (F1InstanceLoss). Some approaches exist for ECG delineation \cite{Jimenez-Perez2019, Jimenez-Perez2021, Tison2019, Camps2019, Hedayat2018} but, to the best of our knowledge, no approaches exist in the literature that combine a quantification task through explicit (rule-based synthetic data generation) and implicit (application of the BoundaryLoss and F1InstanceLoss functions) prior imposition.

Performance-wise, the developed models compare favorably with existing DSP-based and ML-based approaches found in the literature. Our best performing model surpassed the state-of-the-art in the usual precision and delineation metrics. We obtained an average $F_1$ score of 99.38\% and onset and offset errors of $2.19 \pm 17.73$ ms and $4.45 \pm 18.32$ ms with respect to the reference for all waves in the QT database, as detailed in Table \ref{tab:resultsbest}. These figures represent a precision gain of up to 8\% at some fiducials such as the P wave, recall values nearing 100\% and an overall reduced SD, surpassing the best-performing methods in most categories. Some delineation metrics, however, present decreased performance as compared to DSP-based methods, which might be explained by their higher control on the decision boundaries that produce a certain prediction, to the difficulties at localizing onsets and offsets in smoothly increasing/decreasing waves such as the P and T waves and to the fact that, at the sampling rate of the predictions ($250$ Hz), a $4$ ms difference is equivalent to a single sample.

\begin{figure*}[!p]
    \centering
    \includegraphics[width=1\linewidth]{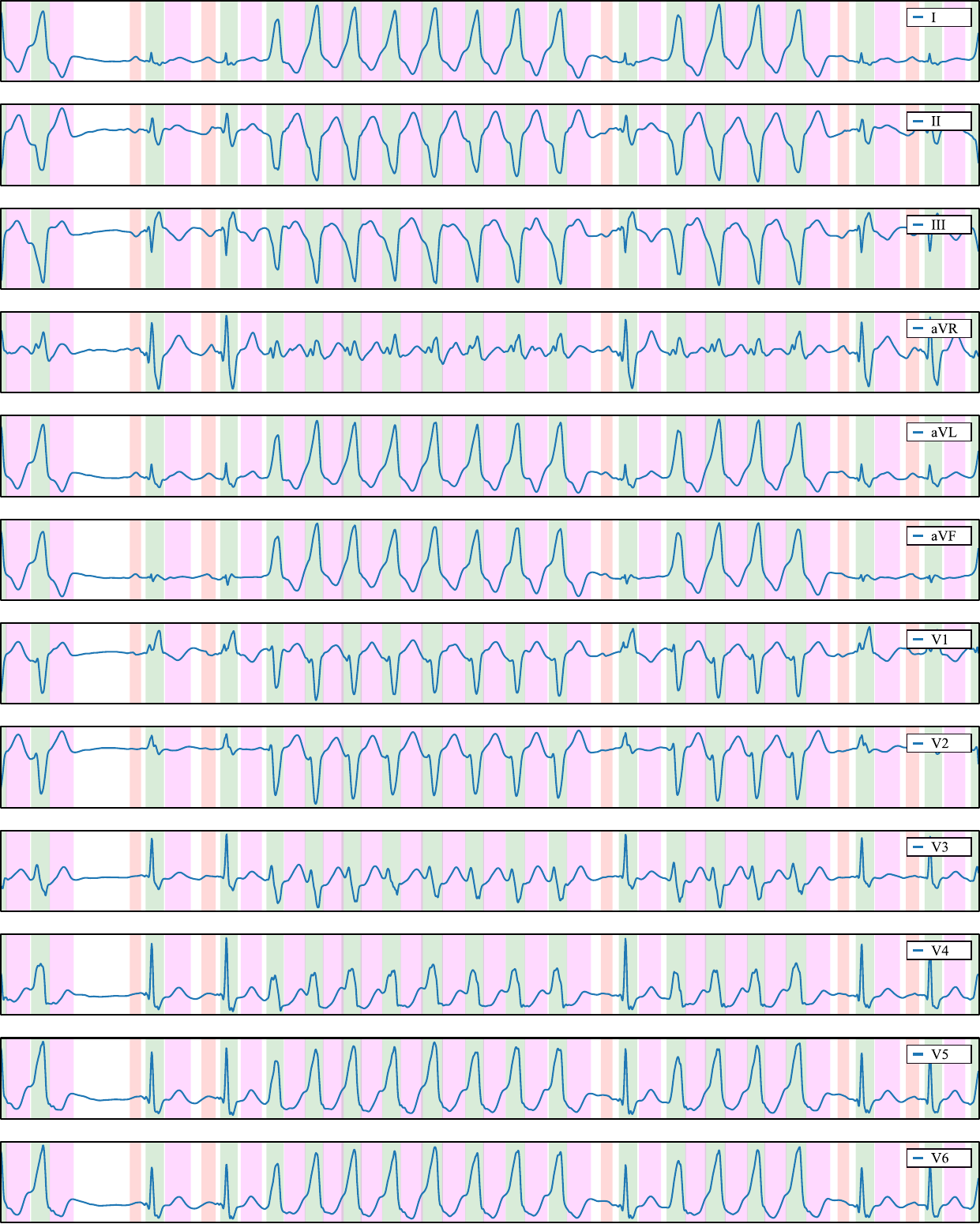}
    \caption{\label{fig:zhejiangpredictionTV} Delineation prediction of the sample ``922551'' of the Zhejiang database, containing a non-sustained ventricular tachycardia. Color code: P wave (red), QRS wave (green) and T wave (magenta).
}
\vspace{2em}
\end{figure*}

Secondly, the trained model presented good generalization properties when predicting samples from the QT database (Table \ref{tab:resultsbest}) as well as in the LU \cite{kalyakulina2020ludb} and the Zhejiang \cite{zheng202012} databases (Table \ref{tab:resultsluzhejiang}). Specifically, the performance on the LU database surpasses all approaches in the literature and even displays $F_1$-scores of 100\% for the QRS and T waves, which hint to both the robustness of our model and at the relative simplicity of the represented rhythms in the database. The performance on the more challenging Zhejiang database does not reflect reduced performance, providing metrics that closely resemble those of the QT database. A prediction from the Zhejiang database is depicted in Figure \ref{fig:zhejiangpredictionTV}.

As presented in Section \ref{sec:bestadditions}, the increased model performance can be directly related to the adopted design decisions, i.e., the addition of pseudo-synthetic data generation, the usage of the W-Net architecture and the implementation of the F1InstanceLoss and the BoundaryLoss. The inclusion of synthetic data supposed the best model addition, consistently improving the performance of the model when comparing models trained with and without synthetic generation. Interestingly, models trained exclusively with synthetic data also displayed better metrics than those trained only with real data. The second-to-best model modification was the usage of the W-Net architecture, which can be linked to an increase in model capacity. The third best addition was the usage of novel loss functions. This addition, although it has been shown to improve model performance across all runs, was not present in some of the top models. Further research is needed to assess the mechanism these losses use to increase model performance. The specific $F_1$ and SD figures for all the aforementioned model additions have been reported in Section \ref{sec:bestadditions}.

As it has been discussed, training a delineation model with synthetic data augmentation demonstrates some advantages with respect to the state of the art besides purely enhanced detection and delineation performance. Firstly, it allows for explicit prior imposition in the synthetic data generation process, a philosophy that can be extended to many other quantification domains (e.g., medical images, intracavitary electrograms). Moreover, the pseudo-synthetic generation pipeline addresses the lack of generalization of current existing delineation algorithms, reducing the impact of small databases and low inter-sample variance that has hindered both ML-based and DSP-based approaches \cite{Minchole2019}. An illustrative example is the good performance of the algorithm on the Zhejiang database, which contains rhythms that have been engineered into the synthetic data generation but were not part of the training data. Finally, the predictions produced by the model were much more robust than other state-of-the-art approaches. For instance, some approaches displayed incoherent predictions, where the rise and fall of hyperacute T waves were predicted as two separate waves, completely missed waves, mistagged waves, misplaced onset/offset pairs or noisy activations spanning 1-2 samples \cite{Jimenez-Perez2021,Hedayat2018}.

There are, however, some limitations to the presented approach. Firstly, the set of rules developed in the data generation is too narrow. Many more conditions could be represented, and richer modifications over the fundamental ECG segments (cropped P, QRS, T waves) could be applied, such as addition of delta, J or epsilon waves, or atrial/ventricular hypertrophy. Secondly, the rise in computation time tied to the on-the-fly data generation, alongside the existing computational and temporal constraints that are common in the DL literature when training large models, have limited our ability to provide exhaustive testing on the contributions of each element to the final result, especially due to the large amount of tunable hyperparameters. The synthetic generator has been employed with some hyperparameters that produced visually credible samples, but a rigorous validation is still lacking. Moreover, despite the efforts placed on generating VT records, and despite the success in a large percentage of predictions, the network still has difficulties finding the onsets and offsets of very fast VTs/ventricular flutter. This, however, is to be expected as even trained physicians have difficulties at their delimitation. Finally, the network is slightly sensible to input normalization. Given that the amplitude was normalized for sinus rhythm QRS to take values in the range $[0.5, 1]$ (thus taken larger values for other rhythms, such as extrasystoles), we have opted for normalizing the model's input with the median of a moving average over the signal, with a window of 256 samples. This criterion could also be improved upon.

\subsection{Conclusions}\label{sec:conclusions}

This work addresses some of the main challenges in ML-based clinical data analysis: the uninterpretability of classification-based models, the reduced database size and the imposition of data priors. For this purpose, we developed a DL-based pipeline for the automatic quantification of the electrocardiogram through novel prior imposition strategies in the shape of pseudo-synthetic data generation and shape regularization losses.

The produced network has demonstrated remarkable detection and delineation metrics, as well as good generalization when predicting a variety of samples of different open source databases. This allows its application to a large number of downstream tasks, allowing the production of automatic and objective metrics over clinical data, thus becoming an enabling technology for further automatization of ECG analysis. It, however, presents some limitations. Firstly, the synthetic data generation produces a dependence on input data normalization when predicting input samples, although windowing and normalizing to the median usually performs well. Secondly, a larger array of cardiac conditions in the pseudo-synthetic data generation algorithm and a more in-depth exploration of the generative parameters should be performed. Finally, a more exhaustive testing of the performance gain of each model addition could be explored.

\section*{Conflict of Interest Statement}

The authors declare that the research was conducted in the absence of any commercial or financial relationships that could be construed as a potential conflict of interest.

\section*{Author Contributions}

G.J. conceived and conducted the experiments. J.A. reviewed the delineation annotations. G.J., A.A. and O.C. analysed the results. All authors reviewed the manuscript.

\section*{Funding}

This research was supported by the Secretariat for Universities and Research of the Ministry of Business and Knowledge of the Government of Catalonia (2017 FI\_B 01008), by the Gobierno de Aragón and FEDER 2014-2020 (T39\_20R) and by the Maria de Maeztu Units of Excellence Programme (MDM-2015-0502). The GPU was donated by the NVIDIA Corporation.

\subsection*{Data Availability Statement}

The databases used in this work are publicly available at \url{https://physionet.org/content/qtdb/1.0.0/} (QT database), at \url{http://www.cyberheart.unn.ru/database} (LU database) and at \url{https://doi.org/10.6084/m9.figshare.c.4668086.v2} (Zhejiang Hospital database). The ground truth revisions in the QT database and the annotations over the Zhejiang database have been uploaded to Figshare in \cite{figshare_QTDB}, \cite{figshare_LUDB} and \cite{figshare_Zhejiang}. The tool employed for producing the revised ground truth annotations used the Bokeh python library \cite{Bokeh} and is available in the author's repository at \url{https://github.com/guillermo-jimenez/QRSAnnotator}.

\bibliographystyle{unsrt}
\bibliography{refs}

\end{document}